%% file: main.tex
\documentclass{article}

\usepackage[final]{infocog_neurips_2022}

\usepackage[utf8]{inputenc} 
\usepackage[T1]{fontenc}    
\usepackage{url}            
\usepackage{booktabs}       
\usepackage{amsfonts}       
\usepackage{nicefrac}       
\usepackage{microtype}      
\usepackage[usenames, dvipsnames]{color} 
\usepackage{enumerate, amsmath, amsthm, amssymb, algorithm, algorithmic, pifont, csquotes, dashrule, tikz, bbm, booktabs, bm, verbatim, url, dsfont}
\usepackage[framemethod=TikZ]{mdframed}
\usepackage{natbib}
\usepackage[colorlinks]{hyperref}
\hypersetup{
     colorlinks = true,
     linkcolor = blue,
     anchorcolor = blue,
     citecolor = red,
     filecolor = blue,
     urlcolor = blue
     }
\usepackage{caption}
\usepackage{subcaption}
\input{commands}

\newif\ifsubmit
\submittrue
\ifsubmit
\newcommand{\dnote}[1]{}
\newcommand{\mnote}[1]{}
\newcommand{\nnote}[1]{}
\newcommand{\bnote}[1]{}
\else
\newcommand{\dnote}[1]{\textcolor{blue}{Dilip: #1}}
\newcommand{\mnote}[1]{\textcolor{violet}{Mark: #1}}
\newcommand{\nnote}[1]{\textcolor{green}{Noah: #1}}
\newcommand{\bnote}[1]{\textcolor{orange}{Ben: #1}}
\fi

\title{On Rate-Distortion Theory in Capacity-Limited Cognition \& Reinforcement Learning}

\author{%
  Dilip Arumugam\\
  Department of Computer Science\\
  Stanford University\\
  \texttt{dilip@cs.stanford.edu}\\
  \And
  Mark K. Ho\\
  Department of Computer Science\\
  Princeton University\\
  \texttt{mho@princeton.edu}\\
  \And
  Noah D. Goodman \\
  Department of Psychology \\
  Department of Computer Science\\
  Stanford University\\
  \texttt{ngoodman@stanford.edu} \\
  \And
  Benjamin Van Roy \\
  Department of Electrical Engineering \\
  Department of Management Science \& Engineering\\
  Stanford University\\
  \texttt{bvr@stanford.edu} \\
}

\begin{document}

\maketitle

\begin{abstract}
  Throughout the cognitive-science literature, there is widespread agreement that decision-making agents operating in the real world do so under limited information-processing capabilities and without access to unbounded cognitive or computational resources. Prior work has drawn inspiration from this fact and leveraged an information-theoretic model of such behaviors or policies as communication channels operating under a bounded rate constraint. Meanwhile, a parallel line of work also capitalizes on the same principles from rate-distortion theory to formalize capacity-limited decision making through the notion of a \textit{learning target}, which facilitates Bayesian regret bounds for provably-efficient learning algorithms. In this paper, we aim to elucidate this latter perspective by presenting a brief survey of these information-theoretic models of capacity-limited decision making in biological and artificial agents.
\end{abstract}

\section{Motivation}

In a perfect world, decision makers of any shape or form would always be capable of producing optimal behaviors. In reality, however, the inescapable constraints of an overwhelmingly-complex environment force agents to seek out alternative behaviors that are sufficiently satisfying or, more succinctly, satisficing. While satisficing is a longstanding, well-studied idea about how to understand resource-limited cognition~\citep{simon1955behavioral,simon1956rational,newell1958elements,newell1972human,simon1982models}, it has been usually treated as contrary to rational analysis~\citep{anderson1990adaptive}. In particular, modern resource-rational analyses, both utility-theoretic~\citep{griffiths2015rational,lieder2020resource} and information-theoretic~\citep{sims2003implications,sims2016rate,gershman2020origin}, still aim to find an optimal policy but to do so within resource constraints. An alternative view for the resource-rational learning of satisficing agents comes from recent theoretical work in bandit learning and reinforcement learning~\citep{arumugam2021deciding,arumugam2022deciding}; this perspective focuses on learning to achieve a deliberately-sub-optimal, satisficing policy that requires obtaining fewer bits of information from the environment. Unlike other approaches to resource-rational analysis, this methodology is accompanied by guarantees about provably-efficient learning through its use of epistemic uncertainty~\citep{der2009aleatory} to resolve the underlying exploration-exploitation trade-off. 

\section{Rate-Distortion Theory}

Here we offer a brief, high-level overview of rate-distortion theory~\citep{shannon1959coding,berger1971rate,cover2012elements}. Due to space constraints, precise mathematical details are relegated to Appendix \ref{sec:prelims}. A lossy compression problem consumes as input a fixed information source $\bP(X \in \cdot)$ and a distortion function $d: \mc{X} \times \mc{Z} \ra \bR_{\geq 0}$ which quantifies the loss of fidelity by using an element $z \in \mc{Z}$ in lieu of $x \in \mc{X}$. Then, for any $D \in \bR_{\geq 0}$, the rate-distortion function quantifies the fundamental limit of lossy compression as $$\mc{R}(D) = \inf\limits_{Z \in \mc{Z}} \bI(X;Z) \triangleq \inf\limits_{Z \in \mc{Z}} \bE\left[\kl{\bP\left(X \in \cdot \mid Z\right)}{\bP(X \in \cdot)}\right] \text{ such that } \bE\left[d(X,Z)\right] \leq D,$$ where the infimum is taken over all random variables $Z$ (representing the output of a channel given $X$ as input) that incur bounded expected distortion, $\bE\left[d(X,Z)\right] \leq D$. As an aside, we note that one may equivalently define $\mc{R}(D)$ as an infimum over a constrained collection of either joint distributions (as done in \citep{csiszar1974extremum,dembo2002source}, for example) or conditional distributions (such as in \citep{blahut1972computation,kawabata1994rate}), representing the channel or lossy compression itself. Naturally, $\mc{R}(D)$ represents the minimum number of bits of information that must be retained on average from $X$ in order to achieve this bound on the expected loss of fidelity. Moreover, $\mc{R}(D)$ is well-defined for arbitrary information source and channel output random variables taking values on abstract spaces~\citep{csiszar1974extremum}. In certain contexts, it can be more suitable to employ the inverse of $\mc{R}(D)$, the distortion-rate function: $\mc{D}(R) = \inf\limits_{Z \in \mc{Z}} \bE\left[d(X,Z)\right] \text{ such that } \bI(X;Z) \leq R.$ For a given upper limit $R \in \bR_{\geq 0}$ on the bits of information that can be transmitted, $\mc{D}(R)$ quantifies the minimum achievable distortion of the resulting compression.

\section{Problem Formulation}

We formulate a sequential decision-making problem as an episodic, finite-horizon Markov Decision Process (MDP)~\citep{bellman1957markovian,Puterman94} defined by $\mc{M} = \langle \mc{S}, \mc{A}, \mc{U}, \mc{T}, \beta, H \rangle$. Here $\mc{S}$ denotes a set of states, $\mc{A}$ is a set of actions, $\mc{U}:\mc{S} \times \mc{A} \ra [0,1]$ is a deterministic reward or utility function providing evaluative feedback signals, $\mc{T}:\mc{S} \times \mc{A} \ra \Delta(\mc{S})$ is a transition function prescribing distributions over next states, $\beta \in \Delta(\mc{S})$ is an initial state distribution, and $H \in \bN$ is the maximum length or horizon. Within each one of $K \in \bN$ episodes, the agent acts for exactly $H$ steps beginning with an initial state $s_1 \sim \beta$. For each timestep $h \in [H]$, the agent observes the current state $s_h \in \mc{S}$, selects action $a_h \sim \pi_h(\cdot \mid s_h) \in \mc{A}$, enjoys a reward $r_h = \mc{U}(s_h,a_h) \in [0,1]$, and transitions to the next state $s_{h+1} \sim \mc{T}(\cdot \mid s_h, a_h) \in \mc{S}$.

A stationary, stochastic policy for timestep $h \in [H]$, $\pi_h:\mc{S} \ra \Delta(\mc{A})$, encodes behavior as a mapping from states to distributions over actions. Letting $\{\mc{S} \ra \Delta(\mc{A})\}$ denote the class of all stationary, stochastic policies, a non-stationary policy $\pi = (\pi_1,\ldots,\pi_H) \in \{\mc{S} \ra \Delta(\mc{A})\}^H$ is a collection of exactly $H$ stationary, stochastic policies whose overall performance in any MDP $\mc{M}$ at timestep $h \in [H]$ when starting at state $s \in \mc{S}$ and taking action $a \in \mc{A}$ is assessed by its associated action-value function $Q^\pi_{\mc{M},h}(s,a) = \bE\left[\sum\limits_{h'=h}^H \mc{U}(s_{h'},a_{h'}) \bigm| s_h = s, a_h = a\right]$, where the expectation integrates over randomness in the action selections and transition dynamics. Taking the corresponding value function as $V^\pi_{\mc{M},h}(s) = \bE_{a \sim \pi_h(\cdot \mid s)}\left[Q^\pi_{\mc{M},h}(s,a)\right]$, we define the optimal policy $\pi^\star = (\pi^\star_1,\pi^\star_2,\ldots,\pi^\star_H)$ as achieving supremal value $V^\star_{\mc{M},h}(s) = \sup\limits_{\pi \in \{\mc{S} \ra \Delta(\mc{A})\}^H} V^\pi_{\mc{M},h}(s)$ for all $s \in \mc{S}$, $h \in [H]$. 
We let $\tau_k = (s^{(k)}_1, a^{(k)}_1, r^{(k)}_1, \ldots,s^{(k)}_{H}, a^{(k)}_{H}, r^{(k)}_{H}, s^{(k)}_{H+1})$ be the random variable denoting the trajectory experienced by the agent in the $k$th episode. Meanwhile, $H_k = \{\tau_1,\tau_2,\ldots, \tau_{k-1}\} \in \mc{H}_k$ is the random variable representing the entire history of the agent's interaction within the environment at the start of the $k$th episode. 
\section{Capacity Limitation as a Policy Information Bottleneck}

There is a long, rich literature exploring the natural limitations on time, knowledge, and cognitive capacity faced by human (and animal) decision makers~\citep{simon1956rational,newell1958elements,newell1972human,simon1982models,gigerenzer1996reasoning,vul2014one,griffiths2015rational,gershman2015computational,icard2015resource,lieder2020resource,bhui2021resource,brown2022humans,ho2022people}. Crucially, our focus is on a recurring theme throughout this literature of modeling these limitations on cognitive capabilities as being information-theoretic in nature~\citep{sims2003implications,peng2005learning,parush2011dopaminergic,botvinick2015reinforcement,sims2016rate,sims2018efficient,zenon2019information,ho2020efficiency,gershman2020reward,gershman2020origin,mikhael2021rational,lai2021policy,gershman2021rational,jakob2022rate,bari2022undermatching}. Broadly speaking, these approaches all center around the perspective that a policy $\pi_h: \mc{S} \ra \Delta(\mc{A})$ should be modeled as a communication channel that, like a human decision-maker with limited information processing capability, is subject to a constraint on the maximal number of bits that may be sent across it. Consequently, an agent aspiring to maximize returns must do so subject to this constraint on policy complexity; conversely, an agent ought to transmit the minimum amount of information possible while it endeavors to reach a desired level of performance~\citep{polani2009information,polani2011informational,tishby2011information,rubin2012trading}. Paralleling the distortion-rate function $\mc{D}(R)$, the resulting policy-optimization objective follows as $\sup\limits_{\pi \in \{\mc{S} \ra \Delta(\mc{A})\}^H} \bE\left[Q^\pi(S, A)\right] \text{ such that } \bI(S; A) \leq R.$ Depending on the precise work, subtle variations on this optimization problem exist from choosing a fixed state distribution for the random variable $S$~\citep{polani2009information,polani2011informational},  incorporating the state visitation distribution of the policy being optimized~\citep{still2012information,gershman2020origin,lai2021policy}, or assuming access to the generative model of the MDP and decomposing the objective across a finite state space~\citep{tishby2011information,rubin2012trading}. In all of these cases, the end empirical result tends to converge by using variants of the classic Blahut-Arimoto algorithm~\citep{blahut1972computation,arimoto1972algorithm} to solve the Lagrangian associated with the constrained optimization~\citep{boyd2004convex} and produce policies that exhibit higher entropy across states under an excessively limited rate $R$, with a gradual convergence towards the greedy optimal policy as $R$ increases. 

The alignment between this optimization problem and that of the distortion-rate function is slightly wrinkled by the non-stationarity of the distortion function (here, $Q^\pi$ is used as an analogue to distortion which changes as the policy or channel does) and, when using the policy visitation distribution for $S$, the non-stationarity of the information source. Despite these slight, subtle mismatches with the core rate-distortion problem, the natural synergy between cognitive and computational decision making~\citep{tenenbaum2011grow,lake2017building} has led to various reinforcement-learning approaches that draw direct inspiration from this line of thinking~\citep{klyubin2005empowerment,ortega2011information,still2012information,ortega2013thermodynamics,shafieepoorfard2016rationally,tiomkin2017unified,lerch2018policy,lerch2019rate,abel2019state}, most notably including parallel connections to work on ``control as inference'' or KL-regularized reinforcement learning~\citep{todorov2007linearly,toussaint2009robot,kappen2012optimal,levine2018reinforcement,ziebart2010modeling,fox2016taming,haarnoja2017reinforcement,haarnoja2018soft,galashov2019information,tirumala2019exploiting}. Nevertheless, despite their empirical successes, such approaches lack principled mechanisms for addressing the exploration challenge~\citep{o2020making}. While exploration is quintessentially studied in the multi-armed bandit setting~\citep{lai1985asymptotically,lattimore2020bandit}, we focus our main discussion on reinforcement learning and defer consideration of bandit learning to Appendix \ref{sec:bandits}. 

Similar to human decision making~\citep{gershman2018deconstructing,schulz2019algorithmic,gershman2019uncertainty}, provably-efficient reinforcement-learning algorithms have historically relied upon one of two possible exploration strategies: optimism in the face of uncertainty~\citep{kearns2002near,brafman2002r,kakade2003sample,auer2009near,bartlett2009regal,strehl2009reinforcement,jaksch2010near,dann2015sample,azar2017minimax,dann2017unifying,jin2018q,zanette2019tighter,dong2022simple} or posterior sampling~\citep{osband2013more,osband2017posterior,agrawal2017optimistic,lu2019information,lu2021reinforcement}. While both paradigms have laid down solid theoretical foundations, a line of work has demonstrated how posterior-sampling methods can be more favorable both in theory and in practice~\citep{osband2013more,osband2016deep,osband2016generalization,osband2017posterior,osband2019deep,dwaracherla2020hypermodels}. In the next section, we outline how these latter posterior-sampling algorithms still lack a consideration for agents acting under limited capacity constraints and demonstrate an alternative utilization of rate-distortion theory to help account for such limitations.

\section{Learning Targets for Capacity-Limited Decision Making}

As is standard in Bayesian reinforcement learning~\citep{bellman1959adaptive,duff2002optimal,ghavamzadeh2015bayesian}, neither the transition function nor the reward function are known to the agent and, consequently, both are treated as random variables. An agent's initial uncertainty in the (unknown) true MDP $\mc{M}^\star = (\mc{U}^\star, \mc{T}^\star)$ is reflected by a prior distribution $\bP(\mc{M}^\star \in \cdot \mid H_1)$. Since the regret is a random variable due to our uncertainty in $\mc{M}^\star$, we integrate over this randomness to arrive at the Bayesian regret over $K$ episodes: $\textsc{BayesRegret}(K, \pi^{(1:K)}) = \bE\left[\textsc{Regret}(K, \pi^{(1)},\ldots,\pi^{(K)}, \mc{M}^\star)\right] = \bE\left[\sum\limits_{k=1}^K \left( V^\star_{\mc{M}^\star,1}(s_1) - V^{\pi^{(k)}}_{\mc{M}^\star, 1}(s_1)\right)\right].$

In the following, we will denote the entropy and conditional entropy conditioned upon a specific realization of an agent's history $H_k$, for some episode $k \in [K]$, as $\bH_k(X) \triangleq \bH(X \mid H_k = H_k)$ and $\bH_k(X \mid Y) \triangleq \bH_k(X \mid Y, H_k = H_k)$, for two arbitrary random variables $X$ and $Y$. This notation will also apply analogously to mutual information: $\bI_k(X;Y) \triangleq \bI(X;Y \mid H_k = H_k) = \bH_k(X) - \bH_k(X \mid Y) = \bH_k(Y) - \bH_k(Y \mid X).$ The dependence on the realization of a random history $H_k$ makes $\bI_k(X;Y)$ a random variable and the usual conditional mutual information arises by integrating over this randomness: $\bE\left[\bI_k(X;Y)\right] = \bI(X;Y \mid H_k).$ Additionally, we will also adopt a similar notation to express a conditional expectation given the random history $H_k$: $\bE_k\left[X\right] \triangleq \bE\left[X|H_k\right].$

A natural starting point for addressing the exploration challenge in a principled manner is via Thompson sampling~\citep{thompson1933likelihood,russo2018tutorial}. The Posterior Sampling for Reinforcement Learning (PSRL)~\citep{strens2000bayesian,osband2013more,osband2014model,abbasi2014bayesian,agrawal2017optimistic,osband2017posterior,lu2019information} algorithm does this by, in each episode $k \in [K]$, sampling a candidate MDP $\mc{M}_k \sim \bP(\mc{M}^\star \in \cdot \mid H_k)$ and executing its optimal policy in the environment $\pi^{(k)} = \pi^\star_{\mc{M}_k}$; notably, such posterior sampling guarantees the hallmark probability-matching principle of Thompson sampling: $\bP(\mc{M}_k = M \mid H_k) = \bP(\mc{M}^\star = M \mid H_k)$, $\forall M \in \mathfrak{M}, k \in [K]$. The resulting trajectory $\tau_k$ leads to a new history $H_{k+1} = H_k \cup \tau_k$ and an updated posterior over the true MDP $\bP(\mc{M}^\star \in \cdot \mid H_{k+1})$. 

We recognize that, for complex environments, pursuit of the exact MDP $\mc{M}^\star$ may be an entirely infeasible goal. A MDP representing control of a real-world, physical system, for example, suggests that learning the associated transition function requires the agent internalize laws of physics and motion with near-perfect accuracy. More formally, identifying $\mc{M}^\star$ demands the agent obtain exactly $\bH_1(\mc{M}^\star)$ bits of information from the environment which, under an uninformative prior, may either be prohibitively large by far exceeding the agent's capacity constraints or be simply impractical under time and resource constraints~\citep{lu2021reinforcement}. Consequently, an agent must embrace a satisficing solution and \citet{arumugam2022between,arumugam2022deciding} employ the following rate-distortion function to identify a suitable lossy compression of the underlying MDP $\widetilde{\mc{M}} \in \mathfrak{M}$ whose information an agent may prioritize as an alternative learning target to $\mc{M}^\star$:
$\mc{R}_k(D) = \inf\limits_{\widetilde{\mc{M}} \in \mathfrak{M}} \bI_k(\mc{M}^\star; \widetilde{\mc{M}}) \text{ such that } \bE_k[d(\mc{M}^\star, \widetilde{\mc{M}})] \leq D.$


Here, the rate-distortion function is indexed by each episode $k \in [K]$ as the agent takes $\bP(\mc{M}^\star \in \cdot \mid H_k)$ for the information source to be compressed, allowing for incremental refinement of the learning target as knowledge of the environment accumulates and is transmitted to the ``next'' agent~\citep{tomasello1993cultural,tomasello1999cultural}. By definition, the $\widetilde{\mc{M}}$ that achieves this rate-distortion limit will demand that the agent acquire fewer bits of information than what is needed to identify $\mc{M}^\star$. Since, the rate-distortion function is a non-negative; convex; and non-increasing function in its argument~\citep{cover2012elements}, the preceding claim is guaranteed for all $k \in [K]$ and any $D > 0$: $\mc{R}_k(D) \leq \mc{R}_k(0) \leq \bI_k(\mc{M}^\star; \mc{M}^\star) = \bH_k(\mc{M}^\star)$.

\citet{arumugam2022deciding} study two distinct choices of distortion function to assess the loss of fidelity incurred by learning a compressed MDP over the original; the first of these provides an information-theoretic account of recent successes in deep model-based reinforcement learning~\citep{silver2017predictron,farahmand2017value,oh2017value,asadi2018lipschitz,farahmand2018iterative,d2020gradient,abachi2020policy,cui2020control,ayoub2020model,schrittwieser2020mastering,nair2020goal,nikishin2022control,voelcker2022value} through the value-equivalence principle~\citep{grimm2020value,grimm2021proper}, where agents deliberately forego learning the true model of the environment in exchange for some approximate surrogate, which discards irrelevant environment features and only models dynamics of the world critical to agent performance. In the interest of space, we focus on the second distortion function which has a simpler form: $$d_{Q^\star}(\mc{M}, \widehat{\mc{M}}) = \sup\limits_{h \in [H]} ||Q^\star_{\mc{M},h} - Q^\star_{\widehat{\mc{M}},h}||_\infty^2 = \sup\limits_{h \in [H]} \sup\limits_{(s,a) \in \mc{S} \times \mc{A}} | Q^\star_{\mc{M},h}(s,a) - Q^\star_{\widehat{\mc{M}},h}(s,a)|^2.$$ Crucially, \citet{arumugam2022deciding} establish an information-theoretic Bayesian regret bound for a posterior-sampling algorithm that performs probability matching with respect to $\widetilde{\mc{M}}$ instead of $\mc{M}^\star$: $\textsc{BayesRegret}(K, \pi^{(1:K)}) \leq \sqrt{\overline{\Gamma}K\mc{R}^{Q^\star}_1(D)} + 2K(H+1)\sqrt{D}.$ Such an algorithm, by virtue of probability matching, explicitly links an agent's exploration strategy not only to its epistemic uncertainty but also to that $\widetilde{\mc{M}}$ which it aspires to learn~\citep{cook2011science}. The bound communicates that an agent with limited capacity must tolerate a higher distortion threshold $D$ and pursue the resulting compressed MDP that bears less fidelity to the original MDP; in exchange, the resulting number of bits needed from the environment to identify such a simplified model of the world is given as $\mc{R}_1^{Q^\star}(D)$ and guaranteed to be less than the entropy of $\mc{M}^\star$\footnote{Here $\overline{\Gamma} < \infty$ is an (assumed) uniform upper bound to the information ratio~\citep{russo2016information,russo2014learning,russo2018learning} that emerges as an artifact of the analysis.}. Additionally, one can express a near-identical result through the associated distortion-rate function that explicitly takes an agent's capacity of only being able to acquire $R \in \bR_{\geq 0}$ bits into account:
$\textsc{BayesRegret}(K, \pi^{(1:K)}) \leq \sqrt{\overline{\Gamma}KR} + 2K(H+1)\sqrt{\mc{D}^{Q^\star}_1(R)}.$

\section{Conclusion}

A number of recent proposals have approached resource-rationality by combining tools from rate-distortion theory with those of sequential decision-making. Here, we have reviewed a parallel line of work that uses related ideas but within the framework of satisficing and Bayesian reinforcement learning. The distinctive feature of this approach is that it precisely characterizes how capacity-limited learners optimally balance \emph{model complexity} and \emph{value distortion} in a manner that gives rise to \emph{learning targets} which reflect rational satisficing--that is, intelligently choosing how sub-optimal to be based on information acquired during learning. A key challenge for future work will be to translate insights about provably-efficient learning algorithms from this literature into plausible models of human decision-making. More broadly, by modulating the exploration of a satisficing agent to identify a fundamentally different learning target than that of an unconstrained agent, these analyses can provide a framework for understanding how information-theoretic considerations not only shape the products of learning (such as perception, memory, and decisions), but also the process of learning itself.



\bibliographystyle{plainnat}
\bibliography{references}


\newpage
\appendix

\section{Information Theory}
\label{sec:prelims}

Here we introduce various concepts in probability theory and information theory used throughout this paper. We encourage readers to consult \citep{cover2012elements,gray2011entropy,polyanskiy2019lecture,duchi21ItLectNotes} for more background. For readers unfamiliar with measure-theoretic probability, we simply note that while one should feel free to use more familiar definitions of the subsequent information-theoretic quantities~\citep{cover2012elements}, the more general formulation of this section allows us to be agnostic as to whether the associated random variables are discrete, continuous, or even some mixture thereof. Consequently, the results described in this work hold regardless of whether the state-action space is discrete or continuous; naturally, however, such details do have consequences for practical instantiations of these ideas.

All random variables are defined on a probability space $(\Omega, \mc{F}, \bP)$. For any natural number $N \in \bN$, we denote the index set as $[N] \triangleq \{1,2,\ldots,N\}$. For any arbitrary set $\mc{X}$, $\Delta(\mc{X})$ denotes the set of all probability distributions with support on $\mc{X}$. For any two arbitrary sets $\mc{X}$ and $\mc{Y}$, we denote the class of all functions mapping from $\mc{X}$ to $\mc{Y}$ as $\{\mc{X} \ra \mc{Y}\} \triangleq \{f \mid f:\mc{X} \ra \mc{Y}\}$.

We define the mutual information between any two random variables $X,Y$ through the Kullback-Leibler (KL) divergence: $$\bI(X;Y) = \kl{\bP((X,Y) \in \cdot)}{\bP(X \in \cdot) \times \bP(Y \in \cdot)} \qquad \kl{P}{Q} = \begin{cases} \int \log\left(\frac{dP}{dQ}\right) dP & P \ll Q \\ +\infty & P \not\ll Q \end{cases},$$ where $P$ and $Q$ are both probability measures on the same measurable space and $\frac{dP}{dQ}$ denotes the Radon-Nikodym derivative of $P$ with respect to $Q$. An analogous definition of conditional mutual information holds through the expected KL-divergence for any three random variables $X,Y,Z$:
$$\bI(X;Y \mid Z) = \bE\left[\kl{\bP((X,Y) \in \cdot \mid Z)}{\bP(X \in \cdot \mid Z) \times \bP(Y \in \cdot \mid Z)}\right].$$
With these definitions in hand, we may define the entropy and conditional entropy for any two random variables $X,Y$ as $$\bH(X) = \bI(X;X) \qquad \bH(Y \mid X) = \bH(Y) - \bI(X;Y).$$ This yields the following identities for mutual information and conditional mutual information for any three arbitrary random variables $X$, $Y$, and $Z$:
$$\bI(X;Y) = \bH(X) - \bH(X \mid Y) = \bH(Y) - \bH(Y | X), \qquad \bI(X;Y|Z) = \bH(X|Z) - \bH(X \mid Y,Z) = \bH(Y|Z) - \bH(Y | X,Z).$$
Through the chain rule of the KL-divergence and the fact that $\kl{P}{P} = 0$ for any probability measure $P$, we obtain another equivalent definition of mutual information, $$\bI(X;Y) = \bE\left[\kl{\bP(Y \in \cdot \mid X)}{\bP(Y \in \cdot)}\right].$$ 

\section{Multi-Armed Bandits}
\label{sec:bandits}

A special case of the finite-horizon episodic MDP formulated earlier is the multi-armed bandit (MAB) problem~\citep{lai1985asymptotically,bubeck2012regret,lattimore2020bandit} which consists of exactly one state $|\mc{S}| = 1$ and a horizon $H = 1$; for simplicity, we will further assume a finite number of arms $|\mc{A}| < \infty$ and follow the standard convention of denoting the total number of time periods as $T = K \in \bN$. Compactly, the MAB is now characterized by an environment $\mc{E} = \langle \mc{A}, \mc{U} \rangle$ where $\mc{U}: \mc{A} \ra [0,1]$ returns the mean reward associated with the input action. While $\mc{A}$ is considered a known quantity, it is the agent's uncertainty in the underlying rewards of the environment $\mc{E}$ that drive its uncertainty in the optimal action $A^\star = \max\limits_{a \in \mc{A}} \mc{U}(a).$ At each time period $t \in [T]$, the agent's initial/refined uncertainties in $\mc{E}$ are given by the prior/posterior distribution $\bP(\mc{E} \in \cdot \mid H_t)$ and it aims to minimize the Bayesian regret: $\textsc{BayesRegret}(T, \pi^{(1:T)}) = \bE\left[\sum\limits_{t=1}^T \left( \mc{U}(A^\star) - \mc{U}(A_t)\right)\right].$

A standard algorithm for solving MABs is Thompson sampling~\citep{thompson1933likelihood,russo2018tutorial} where, at each time period, the agent draws a single posterior sample from its current beliefs over the environment $\theta \sim \bP(\mc{E} \in \cdot \mid H_t)$ and then acts optimally with respect to this sample $\min\limits_{\pi \in \Delta(\mc{A})} \bE\left[\mc{U}(A^\star) - \mc{U}(A_t) \mid \mc{E} = \theta\right]$. Such an action selection strategy embodies the hallmark probability-matching principle of Thompson sampling where, for all $t \in [T]$ and $a \in \mc{A}$, actions are chosen according to their probability of being optimal: $\bP(A_t = a \mid H_t) = \bP(A^\star = a \mid H_t).$ Moreover, this algorithm yields provably-efficient exploration via an information-theoretic Bayesian regret bound~\citep{russo2016information} in terms of the agent's prior entropy over $A^\star$: $\textsc{BayesRegret}(T, \pi^{(1:T)}) \leq \sqrt{\frac{1}{2}|\mc{A}|\bH_1(A^\star)T}$. Since the optimal action $A^\star$ is a deterministic function of the environment $\mc{E}$, $\bH_1(A^\star \mid \mc{E}) = 0$ and so the total amount of information an agent must acquire from the environment $\mc{E}$ over the course of learning to identify $A^\star$ is given by $\bI_1(\mc{E}; A^\star) = \bH_1(A^\star) - \bH_1(A^\star \mid \mc{E}) = \bH_1(A^\star)$. Consequently, the regret bound affirms that an agent who begins with a strong, well-specified prior over $A^\star$ at the start of learning incurs far less regret than an agent operating under a less informative prior. 

Critically, however, decision-making agents operating under constraints on time and resources may not be capable of acquiring all $\bH_1(A^\star)$ bits of information needed for optimal decisions. This reality coupled with the observation that $A^\star = f(\mc{E})$ is merely a (deterministic) statistic $f(\cdot)$ of the unknown environment $\mc{E}$ motivates the consideration of a learning target $\chi \in \mc{A}$~\citep{lu2021reinforcement} that, at the most abstract level, is merely some other statistic of the environment. Of course, $A^\star$ is a statistic of $\mc{E}$ that has a certain desirable property: no other action can achieve mean reward higher than $A^\star$. What characteristic(s) does one want encapsulated in a learning target $\chi$? Intuitively, a good learning target should likely strike some kind of balance between two desiderata: \textbf{(1)} be easier to learn than $A_\star$ by having $\bI_1(\mc{E}; \chi) \leq \bI_1(\mc{E}; A^\star) = \bH_1(A^\star)$ and \textbf{(2)} have bounded performance shortfall or regret relative to $A^\star$, $\bE\left[\mc{U}(A_\star) - \mc{U}(\chi)\right]$.

To this end, \citet{russo2017time,russo2018satisficing,russo2022satisficing} consider a learning target, parameterized by $\eps > 0$, as $A_\eps = \text{Uniform}(\{a \in \mc{A} \mid \mc{U}(A^\star) - \mc{U}(a) \leq \eps\})$ and introduce satisficing Thompson sampling (STS) where probability matching occurs with respect to $A_\eps$: $\bP(A_t = a \mid H_t) = \bP(A_\eps = a \mid H_t)$, $\forall t \in [T], a \in \mc{A}$. While it is intuitive that using an $\eps$-optimal action $A_\eps$ in this manner strikes \textit{some} kind of trade-off between the two desiderata above, it's not clear that this is by any means the \textit{best} trade-off.
Furthermore, while $A^\star$ and $A_\eps$ are chosen and held fixed for the duration of learning, one could imagine adapting a learning target $\chi_t$ incrementally as knowledge of the underlying environment accumulates through $\bP(\mc{E} \in \cdot \mid H_t)$. 

\citet{arumugam2021deciding} leverage the tools of rate-distortion theory to define such a learning target $\tilde{A}_t$ that an agent should strive for in each time period $t \in [T]$ according to the rate-distortion function:
$$\mc{R}_t(D) = \inf\limits_{\tilde{A} \in \mc{A}} \bI_t(\mc{E}; \tilde{A}) \text{ such that } \bE_t\left[d(\tilde{A},\mc{E})\right] \leq D.$$
Assuming that $\mc{E}$ takes values in a set $\Theta$, the distortion function $d_t: \mc{A} \times \Theta \ra \bR_{\geq 0}$ is defined as the expected squared regret of the candidate action: $$d_t(a, \theta) = \bE_t\left[\left(\mc{U}(A^\star) - \mc{U}(a)\right)^2 \mid \mc{E} = \theta\right], \qquad \forall a \in \mc{A}, \theta \in \Theta.$$
Clearly, achieving bounded distortion for any $D \in \bR_{\geq 0}$ addresses the second criterion for learning targets. Since, for all $t \in [T]$ and any $D > 0$, the rate-distortion function $\mc{R}_t(D)$ is a non-negative; convex; and non-increasing function in its argument~\citep{cover2012elements}, it follows that we also ensure the first criterion of a learning target for any $D > 0$: $\mc{R}_t(D) \leq \mc{R}_t(0) \leq \bI_t(\mc{E}; A^\star) = \bH_t(A^\star)$. \citet{arumugam2021deciding,arumugam2021the} employ the classic Blahut-Arimoto algorithm~\citep{blahut1972computation,arimoto1972algorithm} for computing these target actions $\tilde{A}_t$ in each time period and demonstrate a spectrum of regret curves corresponding to various satisficing and optimal policies. Conveniently and unlike with $A^\star$ or $A_\eps$, the onus does not fall upon the agent designer to specify the exact form of the learning target in each time period and, instead, they need only select an appropriate distortion function and threshold for the Blahut-Arimoto algorithm to produce a $\tilde{A}_t$ that strikes the optimal trade-off between ease of learnability and near optimality. \citet{arumugam2021deciding} prove the following information-theoretic Bayesian regret bound for their Blahut-Arimoto Satisficing Thompson Sampling (BLASTS) algorithm: $$\textsc{BayesRegret}(T, \pi^{(1:T)}) \leq \sqrt{2|\mc{A}|\mc{R}_1(D)T} + 2T\sqrt{D},$$ alongside computational experiments demonstrating a wide spectrum of possible target actions that can be realized. Naturally, a setting of $D = 0$ recovers the information-theoretic regret bound of \citet{russo2016information} up to a constant factor. With rewards assumed to be in the unit interval $[0,1]$, a setting of $D = 1$ permits any action to achieve the rate-distortion limit, requiring 0 bits of information overall and rendering the regret bound vacuous. For intermediate settings of $D$, however, the bound critically provides the information-theoretically optimal balance in selecting a learning target that requires the minimum amount of information to deliver a minimum tolerable amount of sub-optimality.

\end{document}

%% file: commands.tex
\definecolor{dblue}{RGB}{98, 140, 190}
\definecolor{dlblue}{RGB}{216, 235, 255}
\definecolor{dgreen}{RGB}{124, 155, 127}
\definecolor{dpink}{RGB}{207, 166, 208}
\definecolor{dyellow}{RGB}{255, 248, 199}
\definecolor{dgray}{RGB}{46, 49, 49}

\newcommand{\durl}[1]{\textcolor{dblue}{\underline{\url{#1}}}}

\newcommand{\eps}{\varepsilon}

\newcommand{\mc}[1]{\mathcal{#1}}

\newcommand{\bE}{\mathbb{E}}

\newcommand{\bR}{\mathbb{R}}
\newcommand{\bP}{\mathbb{P}}

\newcommand{\bI}{\mathbb{I}}
\newcommand{\bH}{\mathbb{H}}
\newcommand{\bN}{\mathbb{N}}

\newcommand{\kl}[2]{D_{\mathrm{KL}}(#1 \mid\mid #2)}

\newcommand{\ra}{\rightarrow}

\newmdenv[
  topline=false,
  bottomline=false,
  rightline = false,
  leftmargin=10pt,
  rightmargin=0pt,
  innertopmargin=0pt,
  innerbottommargin=0pt
]{innerproof}

\newcounter{DaveDefCounter}
